# mAIstro: an open-source multi-agentic system for automated end-to-end development of radiomics and deep learning models for medical imaging

**Research paper**

**Eleftherios Tzanis, PhD[1,*], Michail E. Klontzas, MD, PhD[1,2,3,*]**

1. **Artificial Intelligence and Translational Imaging (ATI) Lab, Department of Radiology, School of Medicine, University of Crete, Heraklion, Crete, Greece**
2. **Computational Biomedicine Laboratory, Institute of Computer Science Foundation for Research and Technology Hellas (ICS - FORTH), Heraklion, Crete, Greece**
3. **Division of Radiology, Department of Clinical Science, Intervention and Technology (CLINTEC), Karolinska Institute, Huddinge, Sweden**

***Corresponding authors**

**Addresses for correspondence**

**Eleftherios Tzanis, PhD**
Postdoctoral Researcher
Artificial Intelligence and Translational Imaging (ATI) Lab
Department of Radiology, School of Medicine,
University of Crete, Voutes, 71003, Heraklion, Crete, Greece
E-mail: tzaniseleftherios@gmail.com
ORCID: 0000-0003-0353-481X

**Michail E. Klontzas, MD, PhD**
Assistant Professor of Radiology
Artificial Intelligence and Translational Imaging (ATI) Lab
Department of Radiology, School of Medicine,
University of Crete, Voutes, 71003, Heraklion, Crete, Greece
Tel: +30 2811391351 E-mail: miklontzas@gmail.com; miklontzas@uoc.gr
ORCID: 0000-0003-2731-933X

**Abstract**

Agentic systems built on large language models (LLMs) offer promising capabilities for automating complex workflows in healthcare AI. We introduce mAIstro, an open-source, fully autonomous multi-agentic framework for end-to-end development and deployment of medical AI models. The system orchestrates exploratory data analysis, radiomic feature extraction, image segmentation, classification, and regression through a natural language interface, requiring no coding from the user. Built on a modular architecture, mAIstro supports both open- and closed-source LLMs, and was evaluated using a large and diverse set of prompts across 16 open-source datasets, covering a wide range of imaging modalities, anatomical regions, and data types. The agents successfully executed all tasks, producing interpretable outputs and validated models. This work presents the first agentic framework capable of unifying data analysis, AI model development, and inference across varied healthcare applications, offering a reproducible and extensible foundation for clinical and research AI integration. The code is available at: https://github.com/eltzanis/mAIstro

## 1. Introduction

The rapid evolution of large language models (LLMs) has influenced multiple domains within healthcare, including clinical support, patient communication and medical education [1, 2]. These models have demonstrated capabilities in translating free-text clinical notes, generating summaries of medical reports, and supporting diagnostic reasoning [3-5]. In radiology and nuclear medicine, LLMs are being tested for report generation, protocol optimization, and structured interpretation assistance [3-6]. Early evaluations suggest that LLMs may improve efficiency and consistency in various medical contexts [6, 7].

However, LLMs lack the ability to autonomously interact with their environment, retrieve and process new datasets, execute code, or directly interface with external pipelines. These limitations restrict their capacity to function as integrated systems in real-world healthcare and biomedical research settings [8, 9].

Agentic systems, particularly those built upon cognitive reasoning architectures, have emerged as a solution to these constraints. These agentic frameworks enable LLMs to reason, plan and execute tasks in iterative cycles of thought, action and observation using external tools. Prompt architectures such as ReAct (Reasoning and Acting) [10], Chain-of-Thought [11] and Tree-of-Thoughts [12] constitute the foundation of such systems, allowing autonomous agents to solve complex problems through structured reasoning steps. These techniques extend the utility of LLMs beyond static text generation, enabling them to perform goal-directed tasks in dynamic environments [13].

The integration of artificial intelligence (AI) in medicine has introduced challenges regarding reproducibility, fairness, transparency and generalizability. Many published AI methodologies and models can not be adopted and implemented in research environments or in routine clinical practice due to poorly standardized development processes or lack of accessible implementation pathways. Moreover, clinicians and researchers without programming skills are excluded from the training, evaluation or deployment of AI tools [14].

To address these challenges, our objective was to develop an open-source, autonomous, multi-agent system capable of understanding and executing natural language instructions for complex biomedical data tasks. The proposed agentic framework is designed to promote accessibility, standardization, and methodological rigor across clinical and research AI workflows. It supports a wide range of tasks such as exploratory data analysis, radiomic feature extraction, image and tabular data modeling, and deployment of state-of-the-art segmentation and classification pipelines, without requiring user interaction beyond natural language prompts.

## 2. Materials and Methods

### a. Multi-agentic system

The agentic system was developed using the smolagents library [15]. The system consists of a master agent, which orchestrates a team of eight task-specific agents. Each agent is designed to perform a specialized task (**Figure 1**). The developed task-specific agents are as follows:

**1. Exploratory Data Analysis (EDA) Agent:** Responsible for performing comprehensive exploratory data analysis, generating descriptive statistics, and visualizations.

**2. Feature Importance and Selection Agent:** Executes various types of feature importance analyses and feature selection methods based on the specific user request.

**3. Radiomic Feature Extraction Agent:** Extracts radiomic features from medical images such as CT and MRI scans.

**4. nnUNet Developer and Implementor Agent:** Automates the training, validation, and deployment of segmentation models using the nnU-Net framework [16] across diverse medical imaging datasets.

**5. TotalSegmentator Agent:** Utilizes the TotalSegmentator framework [17] to automatically segment more than 117 anatomical structures in CT scans and over 50 structures in MRI scans.

**6. Classifier Agent:** Develops, validates, and deploys classification models using tabulated data.

**7. Regressor Agent:** Develops, validates, and deploys regression models based on tabulated data.

**8. Image Classifier Agent:** Trains, evaluates, and implements various image classification models on medical images.

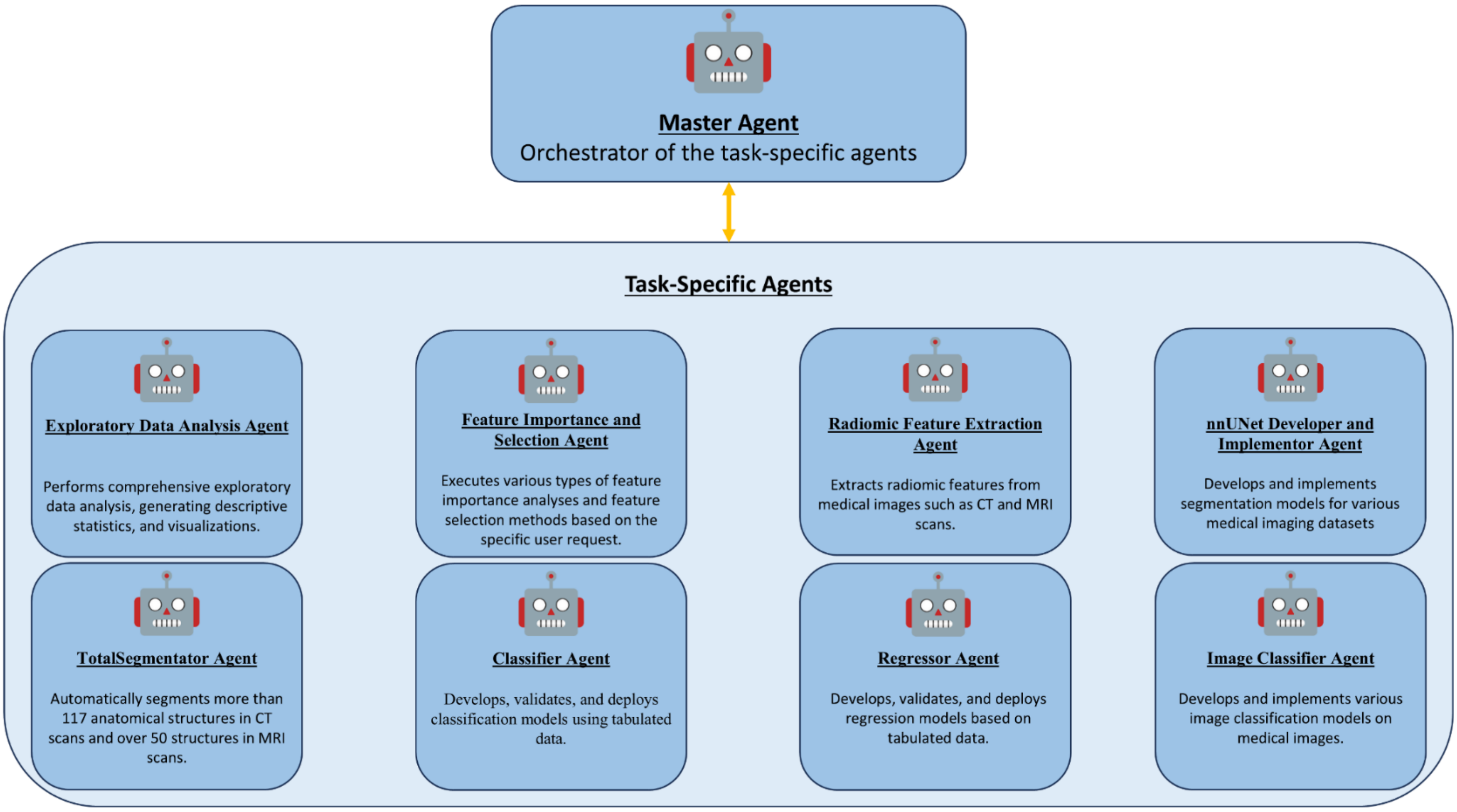

*Figure 1. Architecture of the developed agentic system for autonomous medical data analysis and modeling. The Master Agent orchestrates a team of eight specialized task-specific agents, each designed to autonomously execute distinct functions. Task-specific agents include modules for exploratory data analysis, feature importance analysis, radiomic feature extraction, segmentation model development (nnU-Net and TotalSegmentator), classification and regression model development based on tabulated data, and image classification model training.*

Each agent operates in a cycle of thinking, acting, and observing. Thinking refers to the internal reasoning and planning processes the agent performs to solve the given task. Acting involves interacting with the environment by selecting and executing appropriate actions, such as using tools or gathering information. Observing is the evaluation of action outcomes, allowing the agent to update its internal knowledge and refine its future steps. At its core the agentic system has a Large Language Model (LLM), serving as the "brain" that powers the agent's reasoning and decision-making capabilities. The developed agents are built upon the MultiStepAgent class, a generalized abstraction of the ReAct framework (Reason + Act) introduced by Yao et al. [10].

One of the key components of the proposed system is the design and integration of tools, modular Python classes structured in a standardized way that enables integration with the agents. Each tool includes metadata that allows the agent to understand its purpose, choose it when appropriate, and execute it autonomously to achieve the desired outcome. In total, 16 tools were developed, with specific tools assigned to particular agents based on their respective tasks. In the following sections, the functionality of each task-specific agent, the tools they employ, and the capabilities provided by each tool are described.

**i. Exploratory Data Analysis (EDA) Agent**

The Exploratory Data Analysis (EDA) Agent is responsible for performing a comprehensive exploratory analysis of tabulated datasets, supporting both CSV and Excel file formats. The agent utilizes the ExploratoryDataAnalysisTool class, which provides automated profiling, statistical summarization, visualization, and report generation functionalities.

The tool accepts multiple parameters, including the input file path, the output directory for saving results, optional specifications such as sheet name for Excel files, and settings controlling the visualization and analysis process (e.g., correlation method, sampling strategy, handling of categorical thresholds). Core functionalities of the ExploratoryDataAnalysisTool include:

- Generation of summary statistics and profiling reports.
- Automated creation of visualizations such as histograms, boxplots, pie charts, bar charts, pairplots, correlation heatmaps, and time series analyses.
- Detection and analysis of missing data and outliers.

- Extraction of data structure characteristics including variable types, distribution properties, and relationships with target variables.
- Optional creation of a comprehensive textual summary report.

The ExploratoryDataAnalysisTool operates robustly on large datasets through sampling mechanisms and provides outputs in standardized formats suitable for further machine learning workflows.

**ii. Feature Importance Analysis Agent**

The Feature Importance Analysis Agent is designed to identify the most relevant features for classification or regression tasks based on tabular datasets. It employs the FeatureImportanceAnalysisTool class, which supports multiple feature selection strategies and outputs comprehensive reports. The tool accepts inputs such as the input data path, target column name, task type, and desired feature selection method. Supported methods include random forest importance, ANOVA F-tests, mutual information, and recursive feature elimination (RFE). The tool is capable of both automatic and manual encoding of categorical variables and supports visualization of feature importance and feature distribution. Core functionalities of the FeatureImportanceAnalysisTool include:

- Automated feature selection with configurable strategies.
- Dynamic detection of classification or regression tasks.
- Management of missing values and categorical encoding.
- Generation of CSV files containing selected feature subsets.
- Creation of visual summaries including feature importance plots, cumulative importance curves, principal component analysis (PCA), t-SNE visualizations, and feature correlation heatmaps.

The FeatureImportanceAnalysisTool enables selection of top features at multiple thresholds and is optimized to scale across different dataset sizes and task complexities.

### iii. Radiomics Feature Extraction Agent

The Radiomics Feature Extraction Agent automates the extraction of quantitative radiomic features from medical imaging data. It utilizes the PyRadiomicsFeatureExtractionTool, implemented using the PyRadiomics framework [18].

The tool processes pairs of medical images and corresponding segmentation masks, supporting formats such as NIfTI. It allows detailed configuration of the feature extraction workflow, including selection among multiple image types: Original, Wavelet-filtered, Laplacian of Gaussian (LoG) filtered, Exponential, Gradient, Local Binary Pattern 2D (LBP2D), and Local Binary Pattern 3D (LBP3D) images. Users can specify feature classes to extract, including first-order statistics, shape descriptors, and texture matrices such as GLCM, GLRLM, GLSZM, GLDM, and NGTDM. Specific individual features can also be selected if required.

Preprocessing options include intensity normalization, discretization using a configurable bin width, and optional resampling to isotropic voxel spacing. Feature extraction can be performed either in full 3D or slice-by-slice in 2D mode. The tool supports parallel processing with configurable worker numbers and enables feature extraction targeted to specific labels within the segmentation masks. Associations with external clinical or outcome data can be incorporated through linkage to an external CSV file containing subject-level targets.

The output includes structured CSV files for each label with the extracted features, along with parameter configuration logs and detailed processing reports. This tool enables reproducible, scalable, and standardized generation of high-dimensional radiomic datasets suitable for advanced predictive modeling tasks in clinical imaging research.

### iv. nnU-Net Developer and Implementer Agent

The nnU-Net Developer and Implementer Agent automates the full pipeline for medical image segmentation tasks using the nnU-Net framework. It utilizes two specialized tools: NNUNetTrainingTool and NNUNetInferenceTool.

The NNUNetTrainingTool manages the training process by first preprocessing the input datasets and then training segmentation models across different nnU-Net configurations (2D, 3D full resolution, 3D low resolution and 3D cascade full resolution). The tool supports custom trainers, plan identifiers, transfer learning from pretrained models, multi-GPU training, and flexible control over checkpointing and validation settings. It outputs trained model files and performance metrics such as mean Dice scores, Intersection over Union (IoU), and validation loss.

The NNUNetInferenceTool performs inference by applying trained models to new medical imaging data, generating segmentation masks in standard formats (e.g., NIfTI). It offers control over inference parameters, including device selection, use of test-time augmentation, parallelization options, and checkpoint management.

These tools enable end-to-end automated development, validation, and deployment of deep learning segmentation models within the agentic framework, supporting a wide range of clinical and research imaging applications.

### v. TotalSegmentator Agent

The TotalSegmentator Agent automates the segmentation of anatomical structures in CT and MRI images. It employs the TotalSegmentatorTool class, which interfaces with the TotalSegmentator framework [17] to provide high-resolution multi-organ segmentation. The tool supports input in either NIfTI or DICOM format and allows the selection of predefined segmentation tasks optimized separately for CT and MR imaging. Available options include full-body segmentation, specific organ systems, pathological findings, and radiomics feature extraction.

The TotalSegmentatorTool enables automated segmentation of over 117 structures in CT images and more than 50 structures in MR images, supporting both multilabel outputs and individual binary masks. It allows export of segmentations in NIfTI or DICOM formats, calculation of volumetric statistics, and extraction of radiomic features when required. The tool supports task-specific models, multi-threaded resampling, fast inference modes, device selection (CPU, GPU, or MPS), and memory optimization through chunked processing.

**vi. Classifier Agent**

The Classifier Agent is responsible for the development, evaluation, and deployment of classification models based on tabular data. It utilizes the PyCaret framework [19] through two custom tools: PyCaretClassificationTool for model training and evaluation, and PyCaretInferenceTool for inference on new datasets.

The PyCaretClassificationTool automates the training and comparison of multiple machine learning classification models, leveraging PyCaret's capabilities for hyperparameter tuning, model blending, and dimensionality reduction. It allows control over cross-validation folds, handling of class imbalance, feature preprocessing, GPU acceleration where available, and flexible inclusion or exclusion of specific models. The tool outputs trained models, evaluation metrics, interpretability plots, and structured summary reports.

The PyCaretInferenceTool applies saved models to new datasets, providing prediction outputs and calculating performance metrics when ground truth labels are available. It supports automatic model format handling, detailed evaluation of classification performance for both binary and multiclass problems, and flexible output management.

Together, these tools enable the Classifier Agent to perform automated, reproducible, and scalable classification workflows.

**vii. Regressor Agent**

The Regressor Agent is responsible for the development, evaluation, and deployment of regression models based on tabular data. It utilizes the PyCaret framework through two specialized tools: PyCaretRegressionTool for regression model training and evaluation, and PyCaretRegressionInferenceTool for inference on unseen datasets.

The PyCaretRegressionTool automates the development of predictive models by comparing multiple regression algorithms, performing hyperparameter tuning, creating blended ensembles, and supporting dimensionality reduction techniques. It offers control over preprocessing, feature engineering, GPU utilization, and model selection techniques. Interpretability outputs, including diagnostic plots and feature importance analyses, can also be generated.

The PyCaretRegressionInferenceTool applies trained regression models to external datasets, delivering predictive outputs and quantitative evaluations when ground truth values are available. Performance metrics computed include Mean Squared Error (MSE), Mean Absolute Error (MAE), Root Mean Squared Error (RMSE), $R^2$ score, and Mean Absolute Percentage Error (MAPE), alongside detailed residual analyses.

This combination enables the Regressor Agent to construct optimized regression pipelines, facilitating efficient deployment of high-performing regression models.

### viii. Image Classifier Agent

The Image Classifier Agent is responsible for the development, evaluation, and deployment of convolutional neural network (CNN) models for medical image classification tasks. It utilizes six specialized tools, each implemented using the PyTorch framework [20]: PyTorchResNetTrainingTool, PyTorchResNetInferenceTool, PyTorchVGG16TrainingTool, PyTorchVGG16InferenceTool, PyTorchInceptionV3TrainingTool, and PyTorchInceptionV3InferenceTool.

The PyTorchResNetTrainingTool supports the training and fine-tuning of ResNet architectures [21] (ResNet18, ResNet34, ResNet50, ResNet101, and ResNet152). It allows the optional use of pretrained ImageNet weights or training from scratch, with configurable parameters including the number of epochs, batch size, early stopping with patience control, and dynamic learning rate adjustment. Data augmentation techniques and standardized normalization are applied during training. Model performance is monitored continuously, and checkpoints of the best-performing models are automatically saved.

The PyTorchVGG16TrainingTool provides similar functionality for the VGG16 architecture [22]. It supports both fine-tuning of pretrained VGG16 models and training from randomly initialized weights. Users can configure hyperparameters such as learning rates, number of training epochs, batch sizes, early stopping settings, and data augmentation techniques.

The PyTorchInceptionV3TrainingTool manages the training of InceptionV3 models [23], incorporating specific requirements such as input resizing to 299×299 pixels and the use of auxiliary logits to improve training stability. It also supports pretrained initialization, hyperparameter tuning, early stopping, and multi-stage optimization.

The corresponding PyTorchResNetInferenceTool, PyTorchVGG16InferenceTool, and PyTorchInceptionV3InferenceTool enable batch inference using trained models. These tools provide automated prediction pipelines, outputting class probabilities and predicted labels. They optionally compute classification performance metrics (e.g., accuracy, precision, recall, F1-score, AUC) when ground truth annotations are available and generate evaluation plots such as confusion matrices and ROC curves.

All training and inference tools produce structured output files, including saved models, logs, configuration summaries, evaluation reports, and visualization files, facilitating transparent, reproducible, and scalable deployment of CNN-based classification workflows.

**b. Datasets**

**Tabulated Datasets**

To evaluate the efficiency and versatility of the developed agentic framework, we utilized a series of publicly available tabulated datasets covering both classification and regression tasks:

**i. Breast Cancer Wisconsin (Diagnostic) Dataset:** This dataset [24] contains 569 instances and 30 features derived from digitized images of fine needle aspirates of breast masses. The task is to classify tumors as malignant or benign.

**ii. Heart Failure Clinical Records Dataset:** This dataset [25] contains 299 patient records with 12 clinical features related to heart function and comorbidities, aiming to predict mortality events during follow-up.

**iii. Pima Indians Diabetes Database:** This dataset [26] comprises 768 instances with eight medical predictor variables such as glucose concentration, BMI, and age, aiming to predict the binary presence or absence of diabetes among Pima Indian women.

**iv. Heart Disease Dataset:** This dataset [27] combines data from four sources (Cleveland, Hungary, Switzerland, and the VA Long Beach) and contains 303 instances using 13 clinical features. The objective is to predict the presence of heart disease, categorized as either absence or presence based on angiographic results.

**v. Life Expectancy (WHO) Dataset:** This dataset [28] includes data from 193 countries over multiple years, containing 22 predictors such as immunization rates, mortality rates, and GDP per capita. The primary task is regression to predict life expectancy in years.

These datasets were used to evaluate the task-specific agents operating on tabular data, namely the EDA Agent, Feature Importance Analysis Agent, Classifier Agent, and Regressor Agent. All datasets were used in their original form without synthetic augmentation. Data preprocessing, missing value imputation, feature scaling, and encoding were autonomously handled by the agentic framework.

**Image Classification Datasets**

For the evaluation and testing of the Image Classifier Agent, a selection of datasets from the MedMNIST collection [29, 30] was employed. MedMNIST provides a lightweight benchmark of standardized biomedical imaging datasets for classification tasks. The following datasets were used:

**i. PneumoniaMNIST (28×28) and PneumoniaMNIST (128×128):** Chest X-ray images for binary pneumonia classification.

**ii. PathMNIST (64×64):** Histopathological images for multi-class tissue classification across nine categories.

**iii. BreastMNIST (128×128):** Breast ultrasound images for binary classification of benign versus malignant tumors.

**iv. DermaMNIST (224×224):** Dermoscopy images for seven-class skin disease classification.

**v. OrganAMNIST (28×28):** Abdominal CT images for eleven-class organ classification.

**vi. OCTMNIST (28×28):** Optical coherence tomography images for four-class classification of retinal conditions.

**vii. BloodMNIST (128×128):** Microscopic blood smear images for eight-class hematological cell classification.

All datasets were used with their standard train-validation-test splits as provided. Model training, evaluation, and augmentation were handled autonomously by the agentic system.

**Segmentation and Radiomic Extraction Datasets**

For the evaluation of the Radiomic Feature Extraction Agent, nnU-Net Developer and Implementer Agent, and TotalSegmentator Agent, the following publicly available imaging datasets were utilized:

**i. BraTS 2021 Dataset:** The Brain Tumor Segmentation Challenge 2021 [31] dataset provides multimodal MRI scans (T1, T1Gd, T2, FLAIR) along with expert segmentations of tumor sub regions. It supports the development of segmentation algorithms and radiomic analyses.

**ii. MAMA-MIA Dataset:** The MAMA-MIA dataset [32] offers a large-scale, multi-center collection of breast dynamic contrast-enhanced (DCE) MRI scans with expert tumor segmentations, acquired from 1,506 cases across multiple institutions. It can be used for benchmarking segmentation, radiomics, and clinical prediction modeling.

**iii. KiTS23 Dataset:** The Kidney Tumor Segmentation Challenge 2023 dataset [33] includes high-resolution CT scans with manual segmentations of renal tumors and surrounding structures. It is intended for development and evaluation of automatic segmentation algorithms.

All datasets were utilized in their original format. Preprocessing steps such as resampling, normalization, and intensity standardization, as well as segmentation model development and radiomic feature extraction, were performed autonomously by the agentic framework according to task-specific requirements.

**Experimental Setup**

To evaluate the functionality and robustness of the developed agentic framework, a comprehensive series of queries were executed. These queries were constructed to test two aspects:

i. whether the master agent correctly identifies and deploys the appropriate task-specific agent based on the task requirements described in the query, and

ii. whether the deployed agent successfully performs the assigned task in a fully autonomous and correct manner.

A large and diverse collection of queries was created, covering a wide spectrum of tasks, including exploratory data analysis, feature importance analysis, classification model development and inference, regression model development and inference, radiomic feature extraction, medical image segmentation, and image classification model development and inference.

Specific queries were formulated to test:

**EDA Agent:** Comprehensive exploratory data analysis across multiple tabular datasets.

**Feature Importance Analysis Agent:** Execution of feature selection strategies with varying feature thresholds and datasets.

**Classifier and Regressor Agents:** Development, validation, and inference of machine learning models on structured tabular datasets.

**Radiomic Feature Extraction Agent:** Extraction of radiomic features from CT and MRI scans, using different feature classes and image filters.

**nnU-Net Developer and Implementer Agent:** Training and inference using 3D segmentation models on clinical CT and MRI datasets.

**TotalSegmentator Agent:** Organ-specific and full-body segmentation on CT and MRI images.

**Image Classifier Agent:** Training and inference of deep convolutional neural networks across various MedMNIST datasets using ResNet18, ResNet34, ResNet50, ResNet101, ResNet152, VGG16, and InceptionV3 architectures.

In addition to single-task queries, complex multi-task queries were also constructed to challenge the system's ability to sequentially deploy multiple agents within a single workflow. These included tasks such as segmenting organs, extracting radiomic features from the segmented regions, performing EDA on the extracted features, and developing classification models from the processed datasets.

All experiments were executed in a fully automated manner, with no manual intervention after query submission, thereby validating both the agent selection mechanisms and the task-specific tool execution pipelines of the proposed framework. All tested queries are provided in the Supplementary Material.

To assess the impact of the utilized LLM on the efficiency and reliability of the agentic system, all queries were re-executed using different LLMs as the core reasoning engine. Specifically, the following LLMs were tested: GPT-4o, GPT-4.1, Claude Sonnet 3.7, DeepSeek V3, DeepSeek R1, Llama 3.3 70B, QwQ 32B, Mistral 24B, DeepSeek R1 14B, Llama 4 Scout 17B, Llama 3.1 8B and Mistral 7B. The consistency of agent selection and task execution success rates were recorded for each LLM configuration.

## 3. Results

### System Performance Across Large Language Models

**Table 1** summarizes the success rates across task types and LLMs. A task was considered successful if the master agent correctly identified the required agent, activated it, and the agent autonomously completed the task without intervention or critical errors. Tasks included exploratory data analysis, feature importance analysis, classification model development and inference (tabulated data), regression model development and inference (tabulated data), radiomic feature extraction, medical image segmentation, and image classification model development and inference.

High-performing LLMs, including GPT-4o, GPT-4.1, Claude Sonnet 3.7, DeepSeek V3, DeepSeek R1, and Llama 3.3 70B, achieved a 100% task success rate across all tested tasks. Moderate performance was observed with Llama 4 Scout 17B (91% success rate) and QwQ 32B (90% success rate). In contrast, smaller models such as Llama 3.1 8B, Mistral 24B, Mistral 7B, and DeepSeek R1 14B demonstrated substantially lower success rates, ranging from 10% to 55%.

### Evaluation on Tabulated Data

### EDA and Feature Importance Agents

All queries related to EDA and feature importance analysis were successfully completed across all tested tabulated datasets. The EDA Agent generated comprehensive summary reports, multiple types of visualizations (e.g., histograms, correlation matrices, pairplots, time series plots), and textual insights regarding dataset characteristics. All plots were saved correctly to the specified output directories as requested. The Feature Importance Analysis Agent successfully performed feature ranking. Top-ranked feature sets were saved in structured CSV files as requested, and diagnostic plots were generated accordingly. No errors or incomplete outputs were observed during any EDA or feature selection tasks.

**Classifier Agent**

Blended classification models were developed for four tabulated datasets: Breast Cancer Wisconsin, Heart Disease, Heart Failure, and Pima Indians Diabetes datasets. Each blended model was constructed through the following steps: training of multiple baseline models, selection of the top three models based on cross-validation performance, hyperparameter tuning of the selected models, creation of a final blended ensemble combining the three tuned models. The resulting evaluation metrics, including accuracy, AUC, recall, precision, F1-score, Cohen's kappa, and Matthews correlation coefficient (MCC), are presented in **Table 2**.

**Regressor Agent**

For the regression task on the Life Expectancy dataset, a blended model was similarly developed following model comparison and tuning procedures. The evaluation metrics, including MAE, MSE, RMSE, coefficient of determination ($R^2$), root mean squared log error (RMSLE), and MAPE, are summarized in **Table 3**.

**Evaluation on Medical Image data**

**nnU-Net Developer and Implementer Agent**

The nnU-Net Developer and Implementer Agent was evaluated using the BraTS 2021 (multimodal brain MRI) and KiTS23 (abdominal CT) datasets. For each dataset, a 3D full resolution UNet model was trained and validated using the autonomous agentic workflow.

Evaluation was based on standard segmentation performance metrics, including Dice similarity coefficient (DSC) and Intersection over Union (IoU), calculated on the validation sets. For the BraTS 2021 dataset, whole tumor segmentation achieved a mean DSC of 0.957 and an IoU of 0.920. Tumor core segmentation achieved a mean DSC of 0.951 and an IoU of 0.914. Enhancing tumor segmentation achieved a mean DSC of 0.885 and an IoU of 0.830.

For the KiTS23 dataset, kidney segmentation achieved a mean DSC of 0.951 and an IoU of 0.912 while kidney tumor segmentation achieved a mean DSC of 0.738 and an IoU of 0.637. These results demonstrate that the nnU-Net Agent successfully managed the full training and evaluation workflow for 3D medical image segmentation, achieving high segmentation accuracy on clinically relevant anatomical structures and tumors.

**TotalSegmentator Agent**

The TotalSegmentator Agent was evaluated using images from the BraTS 2021 and KiTS23 datasets. Tasks included full-body segmentation and organ-specific segmentation requests on CT and MRI images. The agent successfully performed segmentation in all queries, generating and saving binary masks for the requested organs and tissues as NIfTI files.

Segmentation outputs were visually inspected and verified for correctness in anatomical location and correspondence to the requested structures. No task failures or incomplete segmentations were observed across the full evaluation set.

**Radiomic Feature Extraction Agent**

The Radiomic Feature Extraction Agent was evaluated on three medical imaging datasets: BraTS 2021 (brain multiparametric MRI), KiTS23 (abdominal CT), and MAMA-MIA (breast DCE-MRI). Radiomic features were extracted across multiple test scenarios, including, extraction from individual segmentation labels (e.g., tumor core, whole kidney), use of different image filters (e.g., original images, wavelet-filtered, Laplacian of Gaussian), selection of specific feature classes (first-order statistics, texture matrices, shape descriptors) and merging extracted radiomic features with additional clinical or demographic predictors when available.

In the case of the MAMA-MIA dataset, the agent successfully merged the extracted radiomic features with external clinical and imaging variables provided by the dataset authors [32].

Across all datasets and scenarios, the agent successfully processed the images and segmentation masks, applied the requested image filters and feature selection, saved structured CSV files containing extracted features, with clear association to each patient ID, and saved parameter configuration logs for reproducibility. Visual inspection and sampling of output files confirmed that all requested features were extracted correctly, merged appropriately with external predictors when specified, and saved in the desired formats and output directories. No task failures were observed during the radiomic extraction experiments.

**Image Classifier Agent**

The Image Classifier Agent was evaluated using a series of datasets from the MedMNIST collection, each corresponding to different imaging modalities and classification challenges. For each dataset, the agent autonomously completed the full pipeline, which included model training, validation and testing based on the structured experimental prompts. Specifically, the agent successfully trained deep neural network architectures (ResNet18, ResNet34, ResNet50, ResNet101, ResNet152, VGG16, and InceptionV3), saved the best-performing model checkpoints as well as the final epoch model, and deployed the trained models for inference on independent test datasets. After inference, the agent evaluated the classification performance using standard metrics, including test set accuracy, macro-averaged precision, macro-averaged recall, and macro-averaged F1-score.

All training and inference tasks were completed without manual intervention or system failures across any dataset. The final evaluation results for each trained model on the respective MedMNIST test sets are summarized in **Table 4**.

**Evaluation of Multi-Agent Sequential Workflows**

To assess the system's ability to handle complex, multi-stage workflows autonomously, a series of multi-task queries were designed and executed. These tasks required the sequential activation and coordination of multiple specialized agents, with intermediate outputs serving as inputs for subsequent tasks.

In the first scenario (multi_task_prompt_1), the TotalSegmentator Agent was deployed to segment the spleen from a set of CT scans. The generated segmentation masks were then used by the Radiomic Feature Extraction Agent to extract spleen-specific radiomic features. Subsequently, the Exploratory Data Analysis Agent performed descriptive analysis of the extracted features, producing statistical summaries **(Figure 2)**. All masks, radiomic features, and EDA outputs were correctly saved to the designated directories, and no failures were observed at any stage.

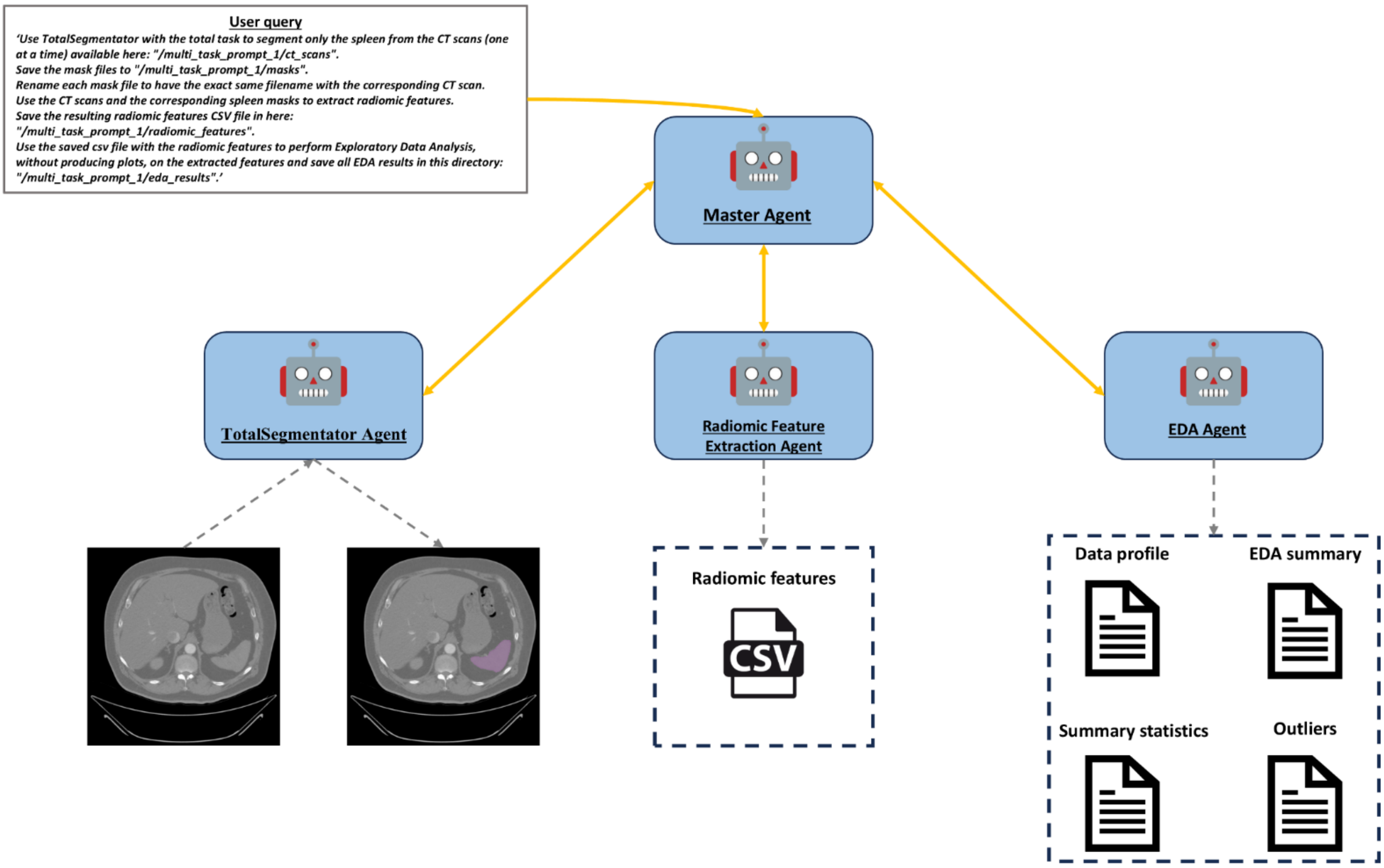

*Figure 2. Workflow execution for multi-task prompt 1, involving a sequence of agent invocations coordinated by the Master Agent. The TotalSegmentator Agent segments the spleen from input CT scans; the resulting masks are used by the Radiomic Feature Extraction Agent to extract quantitative features. These features are then processed by the EDA Agent.*

In the second scenario (multi_task_prompt_2), the nnU-Net Developer and Implementer Agent was activated to segment multiparametric MRI (mpMRI) brain scans from the BraTS 2021 dataset. The Radiomic Feature Extraction Agent subsequently extracted radiomic features from the T1-weighted images across multiple filters (Original, Exponential, Wavelet). Exploratory Data Analysis was performed independently for each tumor subregion, and results were saved in structured subdirectories **(Figure 3)**. All steps were successfully completed without intervention.

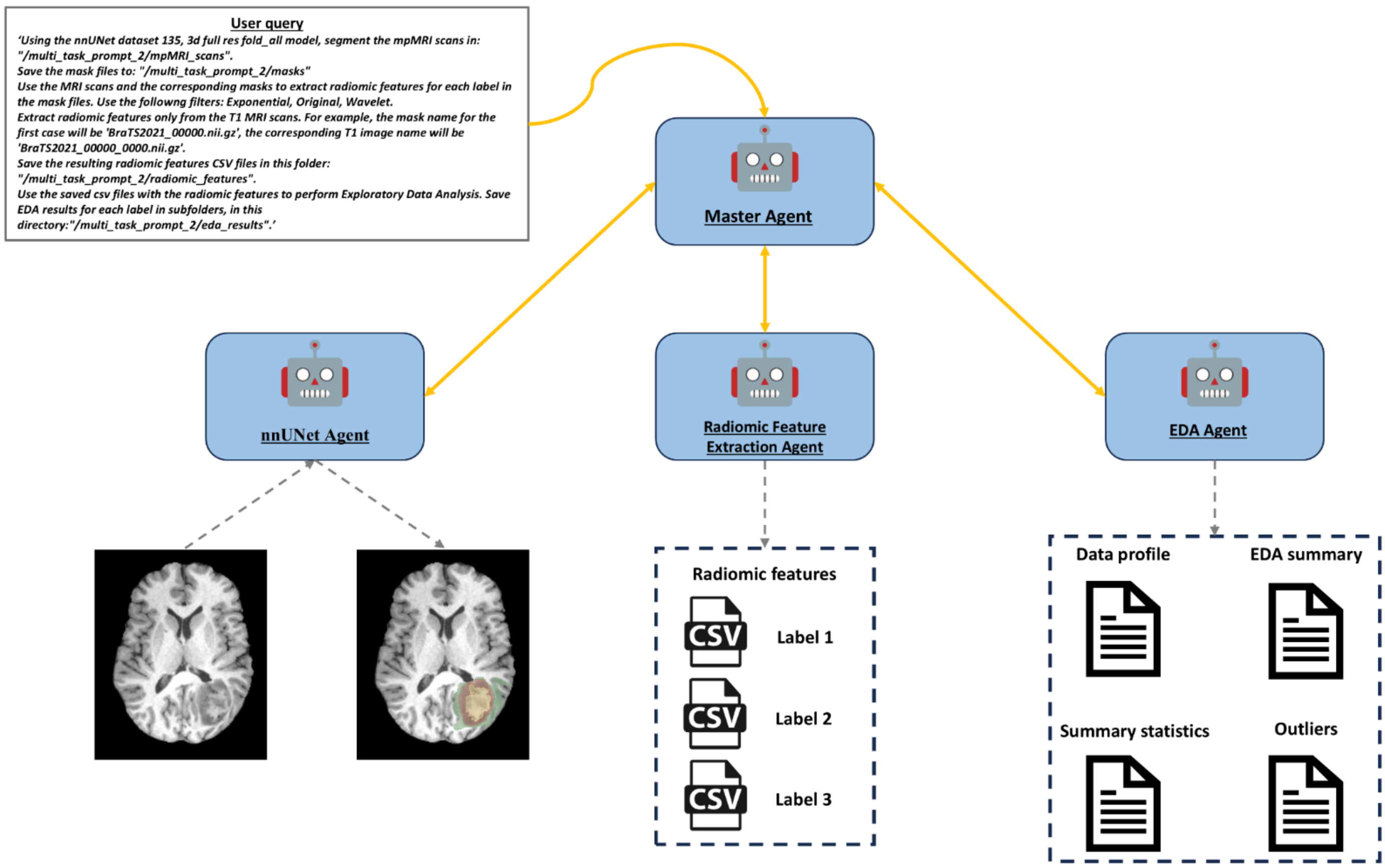

*Figure 3. Workflow for multi-task prompt 2, illustrating autonomous execution of segmentation, radiomics extraction, and data analysis. The nnUNet Agent segments brain tumor subregions from multi-parametric MRI scans using a 3D full-resolution model. Radiomic features are extracted by the Radiomic Feature Extraction Agent from the T1-weighted scans across all labeled tumor regions using multiple filters. The extracted features are then analyzed by the EDA Agent, which generates data summaries for each label.*

The third scenario (multi_task_prompt_3) replicated a clinically realistic research workflow (**Figure 4)** using the newly released MAMA-MIA breast DCE-MRI dataset. In this task, the Feature Importance Analysis Agent first identified the most predictive features for the binary classification target "pcr" (pathological complete response), a clinically important endpoint indicating complete eradication of detectable tumor following neoadjuvant chemotherapy. Subsequently, the Classifier Agent was deployed to train a model using the top 20 features. The resulting blended model achieved a mean cross-validation accuracy of 67.2% (±4.5%), a macro-averaged recall of 77.3% (±7.6%), a macro-averaged precision of 64.6% (±4.7%), and a macro-averaged F1-score of 70.1% (±4.2%). Cohen's kappa and Matthews correlation coefficient were 0.344 and 0.356, respectively. These results demonstrate the system's capability to autonomously process complex clinical datasets and produce clinically meaningful models aligned with current research standards.

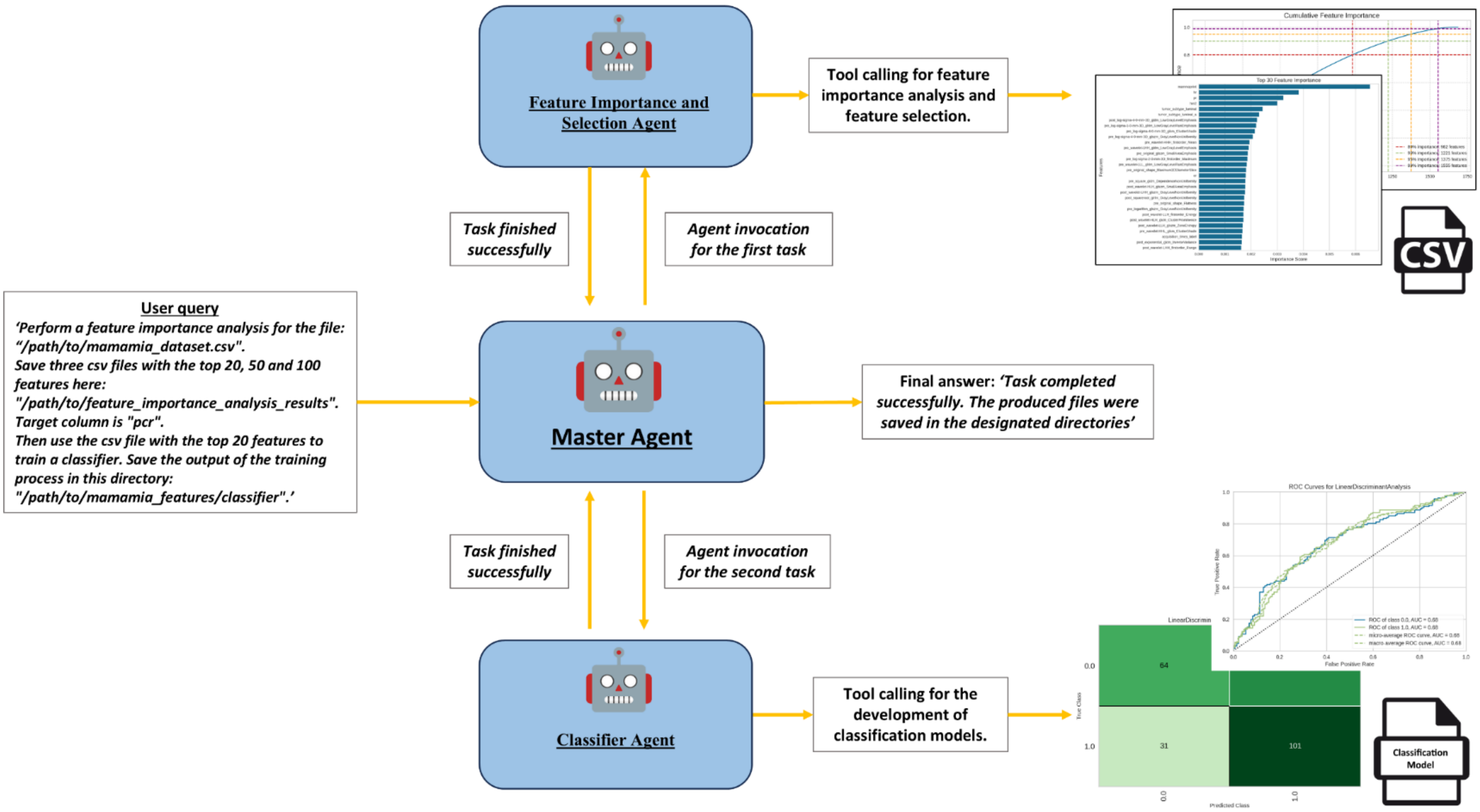

*Figure 4. Multi-agent sequential workflow for real-world clinical data processing based on the MAMA-MIA dataset. A user query requesting sequential feature importance analysis and classifier training is parsed by the Master Agent, which autonomously invokes the appropriate specialized agents. First, the Feature Importance and Selection Agent identifies top-ranked predictive features from the provided CSV file, focusing on the target variable "pcr" (pathological complete response). The resulting features are saved and used to initiate a second task, where the Classifier Agent trains a predictive model. Throughout the workflow, each agent autonomously selects and executes the appropriate tool, saves intermediate outputs and produces final classification results.*

In the fourth scenario (multi_task_prompt_4), a standard tabulated dataset (Breast Cancer Wisconsin) was used. The system performed exploratory data analysis, identified the ten most important features for breast cancer diagnosis, and trained a classification model based on these selected features **(Figure 5)**. All outputs were correctly produced and saved.

The fifth scenario (multi_task_prompt_5) tested a full end-to-end image classification workflow. The Image Classifier Agent successfully trained an InceptionV3 model on the PneumoniaMNIST dataset, achieving a test set accuracy of 80.1%, a macro-averaged precision of 84.9%, a macro-averaged recall of 74.3%, and a macro-averaged F1-score of 75.9%. The area under the ROC curve (AUC) was 0.938. After training, the agent transitioned to perform inference on an independent test set, saving both the trained model and inference results without any manual intervention.

Across all multi-task scenarios, the system demonstrated operational autonomy, correctly managing agent switching, data transfer between tasks, error handling, and organized saving of intermediate and final outputs.

**User query**

*'Perform exploratory data analysis for the file:*
*"/tabulated_datasets/classification/breast_cancer_wisconsin_diagnosis_dataset.csv".*
*Save eda results and a csv files with the top 10 features here: "/test_multi_tasking/multi_task_prompt_4".*
*Targert column is "diagnosis".*
*Then use the csv file with the top 10 features to train a classifier. Save the output of the training process in this directory:*
*"/test_multi_tasking/multi_task_prompt_4/classifier".'*

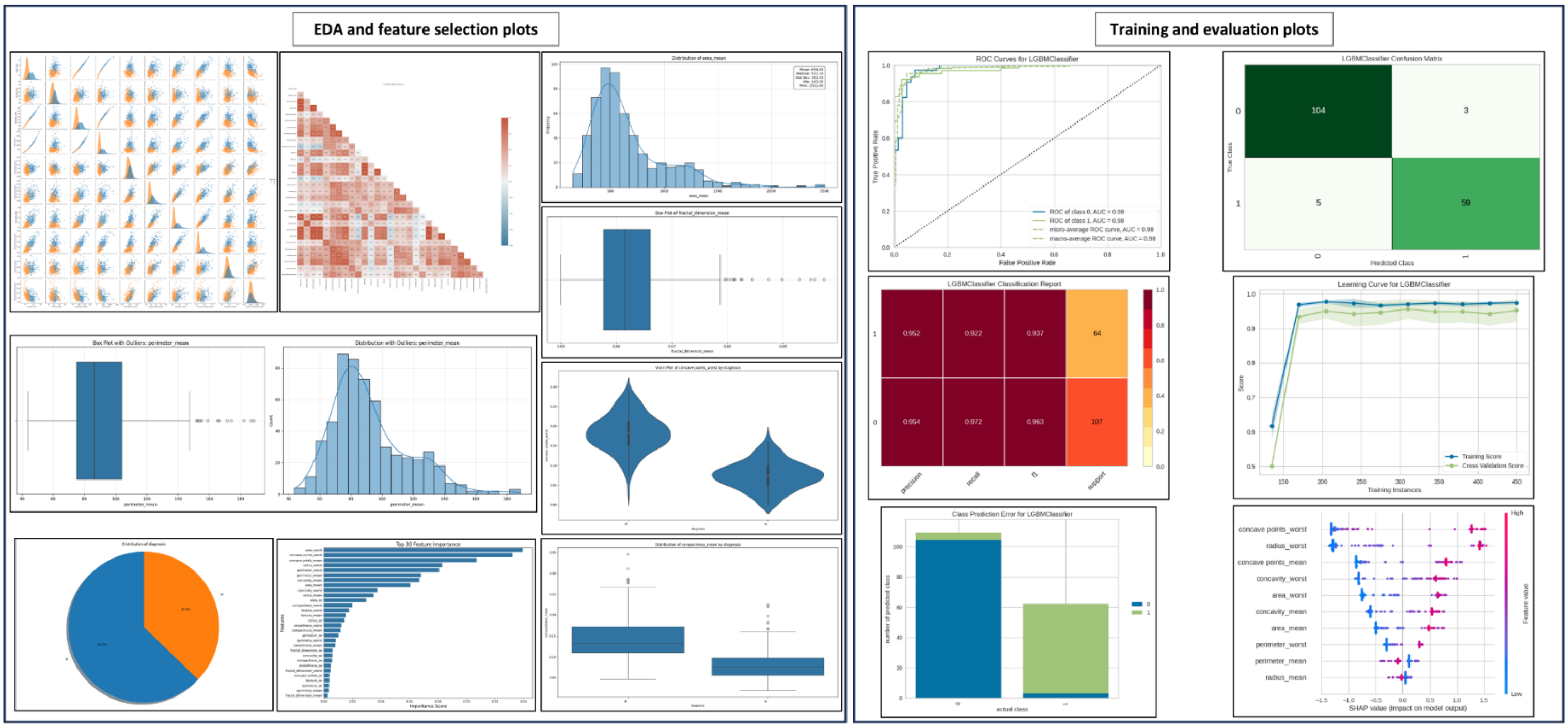

*Figure 5. Illustration of the output generated from multi-task prompt 4, which combined exploratory data analysis, feature selection, and model training on a breast cancer classification dataset. The EDA Agent produced a comprehensive set of visualizations, including correlation heatmaps, pair plots, outlier detection, and class distributions. The right panel displays the resulting model evaluation, including ROC curves, confusion matrix, learning curve, classification metrics, prediction error, and SHAP-based feature interpretation. All outputs were autonomously generated and saved by the agentic system.*

## 4. Discussion

An autonomous multi-agentic system was developed to automate a wide range of medical AI tasks, including EDA, feature importance analysis, radiomic feature extraction, medical image processing, and the development of segmentation and classification models across various imaging modalities. The system was developed to perform these tasks independently and generate comprehensive outputs and evaluation results for each completed process. By releasing mAIstro as an open-source framework, we aim to promote widespread adoption and support the implementation of standardized, reproducible, and well-practiced AI workflows in biomedical research and clinical applications.

The system was evaluated using a diverse set of publicly available datasets comprising both tabulated and medical imaging data. Specifically, five tabular datasets and eleven medical image classification datasets were used, spanning different anatomical regions and clinical applications such as breast cancer diagnosis, heart disease prediction, brain tumor segmentation, pneumonia detection, and more. Evaluation was conducted through a broad set of natural language prompts designed to simulate realistic, task-specific use cases. Furthermore, the framework is LLM-agnostic and was successfully tested with a wide range of proprietary and open-source models, including GPT-4, Claude, DeepSeek, LLaMA, and Qwen. This flexibility allows the system to operate entirely offline in secure environments using local models, when required. Notably, mAIstro enables researchers and clinical personnel without programming knowledge to develop, analyze, and evaluate AI models.

Although prior work has explored the use of LLM agents in healthcare, these studies are typically limited to narrow use cases such as diagnostic decision support or structured clinical simulations. Mehandru et al. [34] proposed a conceptual framework for the use of intelligent agents in clinical environments, emphasizing multi-step reasoning and decision-making, but without implementing or evaluating a functional system. Chen et al. [35] developed a multi-agent conversation (MAC) framework designed for rare disease diagnosis, demonstrating performance improvements through collaborative reasoning between agents, but again limited in scope to diagnostic tasks and structured medical interviews. Another framework, Tang et al.'s MEDAGENTS [36], focused on collaborative question answering across specialties but operated solely through language-based interactions, without the ability to process or analyze real-world data using external tools. To the best of our knowledge, this is the first multi-agentic system designed for end-to-end data handling, EDA, radiomics extraction, model development, evaluation, and deployment across diverse data types in medicine, all orchestrated via natural language and tool-based interaction.

The current study has its limitations. First, the system's performance depends on the reasoning capabilities of the underlying language model, which can vary. Second, tool execution and output interpretation are deterministic, but reasoning remains probabilistic and sensitive to prompt phrasing. Finally, real-world clinical deployment would require further validation under regulatory, ethical, and privacy constraints not addressed herein.

## 5. Conclusion

This study introduces mAIstro, the first open-source multi-agentic framework for autonomous, end-to-end AI development in medical imaging and tabular health data. Capable of performing EDA, radiomic extraction, segmentation, classification, and regression tasks, mAIstro enables non-programmers to interact with and evaluate AI pipelines using natural language. Moreover, experienced users can utilize the framework as a modular foundation—employing the full pipeline or specific agents independently, developing new tools, and extending the system to meet specialized research or clinical needs.

## 6. Data and Code Availability

The code developed for this study is openly available at: https://github.com/eltzanis/mAIstro. All datasets used are publicly accessible through the original sources referenced in the manuscript.

**Acknowledgments**

None.

**Funding**

None.

**Table 1. Query Success Rates Across LLMs and Tasks**

| LLM | Image Classification (Training) | Image Classification (Inference) | Regression (Training) | Regression (Inference) | Classification (Training) | Classification (Inference) | TotalSegmentator | nnUNet | Feature Importance | EDA | Radiomics | Success Rate (%) |
|---|---|---|---|---|---|---|---|---|---|---|---|---|
| GPT-4o | Pass | Pass | Pass | Pass | Pass | Pass | Pass | Pass | Pass | Pass | Pass | 100 |
| GPT-4.1 | Pass | Pass | Pass | Pass | Pass | Pass | Pass | Pass | Pass | Pass | Pass | 100 |
| Sonnet 3.7 | Pass | Pass | Pass | Pass | Pass | Pass | Pass | Pass | Pass | Pass | Pass | 100 |
| DeepSeek V3 | Pass | Pass | Pass | Pass | Pass | Pass | Pass | Pass | Pass | Pass | Pass | 100 |
| DeepSeek R1 | Pass | Pass | Pass | Pass | Pass | Pass | Pass | Pass | Pass | Pass | Pass | 100 |
| Llama 3.3 70B | Pass | Pass | Pass | Pass | Pass | Pass | Pass | Pass | Pass | Pass | Pass | 100 |
| Llama 4 17B | Pass | Pass | Pass | Pass | Pass | Pass | Pass | Fail | Pass | Pass | Pass | 91 |
| Llama 3.1 8B | Fail | Fail | Fail | Fail | Pass | Fail | Fail | Fail | Fail | Pass | Fail | 18 |
| Mistral 24B | Pass | Fail | Fail | Fail | Pass | Fail | Fail | Fail | Pass | Pass | Fail | 36 |
| DeepSeek R1 14B | Pass | Pass | Fail | Fail | Fail | Fail | Pass | Pass | Fail | Pass | Pass | 55 |
| Mistral 7B | Fail | Fail | Fail | Fail | Fail | Fail | Pass | Fail | Fail | Fail | Fail | 10 |
| QwQ 32B | Pass | Pass | Fail | Pass | Pass | Pass | Pass | Pass | Pass | Pass | Pass | 90 |

**Table 2. Classification Models' Results**

| Dataset | Accuracy (%) (SD) | AUC (SD) | Recall (SD) | Precision (SD) | F1-score (SD) | Kappa (SD) | MCC (SD) |
|---|---|---|---|---|---|---|---|
| Breast Cancer | 97.2 (2.1) | 0.998 (0.005) | 97.2 (2.1) | 97.4 (2.0) | 97.2 (2.1) | 0.941 (0.045) | 0.942 (0.044) |
| Heart Disease | 85.4 (5.7) | 0.925 (0.049) | 90.5 (7.1) | 84.6 (7.8) | 87.1 (5.0) | 0.702 (0.117) | 0.714 (0.111) |
| Heart Failure | 73.2 (7.7) | 0.762 (0.085) | 67.6 (24.0) | 59.1 (10.4) | 60.3 (12.8) | 0.411 (0.164) | 0.438 (0.163) |
| Diabetes | 76.2 (6.0) | 0.831 (0.063) | 68.5 (7.7) | 66.2 (9.5) | 66.9 (6.7) | 0.484 (0.117) | 0.488 (0.118) |

**Table 3: Regression Model Results (Life Expectancy Dataset)**

| Metric | Mean (SD) |
|---|---|
| MAE | 0.242 (0.016) |
| MSE | 0.131 (0.025) |
| RMSE | 0.361 (0.035) |
| $R^2$ | 0.9985 (0.0004) |
| RMSLE | 0.0060 (0.0008) |
| MAPE | 0.0038 (0.0003) |

**Table 4. Image Classification Models' Performance**

| Dataset | Model | Accuracy (%) | Macro-averaged Precision | Macro-averaged Recall | Macro-averaged F1-score |
|---|---|---|---|---|---|
| PneumoniaMNIST (28×28) | ResNet18 | 89.1 | 0.919 | 0.857 | 0.876 |
| PathMNIST (64×64) | ResNet34 | 89.7 | 0.909 | 0.873 | 0.877 |
| BreastMNIST (128×128) | ResNet50 | 88.5 | 0.867 | 0.831 | 0.846 |
| DermaMNIST (224×224) | ResNet101 | 79.6 | 0.592 | 0.611 | 0.598 |
| OrganAMNIST (28×28) | ResNet152 | 94.4 | 0.947 | 0.939 | 0.942 |
| OCTMNIST (28×28) | VGG16 | 79.5 | 0.837 | 0.795 | 0.772 |
| BloodMNIST (128×128) | InceptionV3 | 98.9 | 0.991 | 0.990 | 0.991 |

**Supplementary material**

This file contains all natural language prompts used to evaluate the mAIstro system across single-agent and multi-agent tasks.

**Radiomic Feature Extraction from CT**

***Test: Generic query***

rfe_ct_prompt_1 = """

Perform a comprehensive radiomic feature extraction for the CT scans in:

"/path/to/ct/images".

The corresponding masks are here: "/path/to/ct/labels".

Save the results here: "/path/to/output_directory".

"""

***Test: Ask for specific radiomic features and filters.***

rfe_ct_prompt_2 = """

Extract shape and first order radiomic features for the CT scans in: "/path/to/ct/images".

The respective masks are here: "/path/to/ct/labels".

Save the results here: "/path/to/output_directory".

Use the following filters: Exponential, Gradient, LBP2D.

"""

***Test: Generic query for pre contrast MRI images (MAMAMIA dataset)***

rfe_mri_prompt_1 = """

Perform a comprehensive radiomic feature extraction for the MR scans in:

"/path/to/mri/mama_mia/images_pre_contrast".

The respective masks are here: "/path/to//mri/mama_mia/labels".

Save the results here: "/path/to/output_directory".

"""

***Test: Extraction of radiomic features from mpMRI images (BraTS21 dataset) with multiple labels. Asking specific features and filters.***

rfe_mri_prompt_2 = """
Extract shape and glrlm and ngtdm radiomic features for the MR scans in: "/path/to/mri/brats21/images".
The corresponding masks are here: "/path/to/mri/brats21/labels".
Use the followng filters: Exponential, Gradient, SquareRoot.
Save the results here: "/path/to/output_directory".
"""

**Exploratory Data Analysis**

***Prompt for breast wisconsin dataset***

eda_prompt_1 = """

Perform comprehensive exploratory data analysis for the file:

"/path/to/breast_cancer_wisconsin_diagnosis_datasetdata.csv".

Save the output here: "/path/to/output_directory".

"""

***Prompt for predict_diabetes dataset***

eda_prompt_2 ="""

Perform comprehensive EDA for the file: /path/to/predict_diabetes.csv

Save the output here: "/path/to/output_directory"

"""

***Prompt for heart disease dataset***

eda_prompt_3 = """

Perform comprehensive EDA for the file: /path/to/heart_disease_classification.csv.

Save the results here: "/path/to/output_directory".

"""

***Prompt for heart failure dataset***

eda_prompt_4 = """

Perform EDA for the file: /path/to/heart_failure_clinical_records_dataset.csv

Save the results here: "/path/to/output_directory".

"""

***Prompt for life expectancy dataset***

eda_prompt_5 = """

Perform comprehensive EDA for the file: /path/to/Life-Expectancy-Data.csv

Save the results here: "/path/to/output_directory".

"""

**Feature Importance Analysis and Feature Extraction**

***Prompt for breast wisconsin dataset***

fia_prompt_1 = """

Perform feature importance analysis for the file:

"/path/to/breast_cancer_wisconsin_diagnosis_datasetdata.csv".

Save three csv files with the top 5, 10 and 20 features here: "/path/to/output_directory".

Targert column is "diagnosis".

"""

***Prompt for predict_diabetes dataset. Asking to export more features than the original file has.***

fia_prompt_2 = """

Perform feature importance analysis for the file: "/path/to/predict_diabetes.csv".

Save three csv files with the top 5, 10 and 20 features here: "/path/to/output_directory".

Targert column is "Outcome".

"""

***Prompt for heart disease dataset***

fia_prompt_3 = """

Perform feature importance analysis for the file:

"/path/to/heart_disease_classification.csv".

Save three csv files with the top 5, 10 features here: "/path/to/output_directory".

Targert column is "target".

"""

***Prompt for heart failure dataset***

fia_prompt_4 = """

Perform feature importance analysis for the file:

"/path/to/heart_failure_clinical_records_dataset.csv".

Save three csv files with the top 8 features here: "/path/to/output_directory".

Targert column is "DEATH_EVENT".

"""

***Prompt for life expectancy dataset***

fia_prompt_5 = """

Perform feature importance analysis for the file: /path/to/Life-Expectancy-Data-Updated.csv

Save two csv files with the top 10 and 15 features here: "/path/to/output_directory"

Targert column is "Life_expectancy". Create plots.

"""

**nnUNet framework - Train and Inference**

***Prompt for training segmentation UNet (Brats21 dataset)***
nnunet_prompt_1 = 'Train a segmentation 3d full res for the dataset in:
/nnUNet_raw/Dataset135_Brats21. For fold all'

***Prompt for training segmentation UNet (Kits23 dataset)***
nnunet_prompt_1 = 'Train a segmentation 3d full res for the dataset in:
/nnUNet_raw/Dataset140_Kits23. For fold all'

***Prompt for inference mpMRI (Brats21 dataset)***
nnunet_prompt_2 ="""
Using the nnUNet dataset 135, 3d full res fold_all model, segment the scans in:
/inference_nnunet/brats21_validation.
Output folder: /path/to/ouput_directory
"""

***Prompt for inference mpMRI (Kits23 dataset)***
nnunet_prompt_2 ="""
Using the nnUNet dataset 140, 3d full res fold_all model, segment the scans in:
/inference_nnunet/kits23_validation.
Output folder: /path/to/ouput_directory
"""

**TotalSegmentator Inference**

***Prompt 1 segmenting only one organ (CT)***
totalsegmentator_prompt_1 = """
Use TotalSegmentator with the total task to segment only the spleen in the CT scan located at
/path/to/ct_input
Save the mask in this directory: /path/to/ouput_directory
"""

***Prompt 2 segmenting three organs (CT)***
totalsegmentator_prompt_2 = """
Use TotalSegmentator with the total task to segment only the liver, stomach and kidneys in
the CT scan found at /path/to/ct_input
Save the mask here /path/to/ouput_directory
"""

***Prompt 3 segmenting all available organs (CT)***
totalsegmentator_prompt_3 = """
Use TotalSegmentator with the total task to segment all available organs in the CT scan
located at "/path/to/ct_input".
Save the mask in this folder: /path/to/ouput_directory
"""

***Prompt 4 segmenting all available organs (MR)***
totalsegmentator_prompt_4 = """
Use TotalSegmentator with the total_mr task to segment all available organs in the MRI scan
found at "/path/to/ct_input"
Save the mask in this folder: "/path/to/ouput_directory".
"""

**Training prompts Classification Model (Tabulated data)**

***Prompt for classification model development Breast wisconsin***
tct_prompt_1 = """
Train a classification model using the tabulated data:
/path/to/breast_cancer_wisconsin_diagnosis_datasetdata.csv.

Target column: "diagnosis". Exclude lightgbm classifier. Set normalization and transormation to False.
Save the results here: "/path/to/ouput_directory".
"""

***Prompt for classification model development Predict diabetes***

tct_prompt_2 = """
Train a classification model using the file: /path/to/predict_diabetes.csv.
Target column: "Outcome". Exclude lightgbm classifier.
Save the results here: "/path/to/ouput_directory".
"""

***Prompt for classification model development Predict Heart Disease***

tct_prompt_3 = """
Train a classification model using the file: /path/to/heart_disease_classification.csv
Target column: "target". Exclude lightgbm classifier.
Save the results here: "/path/to/ouput_directory".
"""

***Prompt for classification model development Predict Heart Failure***

tct_prompt_4 = """
Train a classification model using the file: /path/to/heart_failure_clinical_records_dataset.csv
Target column: "DEATH_EVENT". Exclude lightgbm, dummy and catboost classifier.
Save the results here: "/path/to/ouput_directory"
"""

**Inference prompts Classification Model (Tabulated data)**

***Prompt for classification inference Breast wisconsin tuned model 1***

ict_prompt_1 = """
Use the classification model:
/path/to/breast_cancer_wisconsin/models/tuned_model_1/tuned_model_1.pkl
Make predictions using the predictors in the file:
/path/to/inferer_results/independent_eval_cohort.csv
The ground truth values are in the column: "diagnosis"
Output directory: "/path/to/ouput_directory"
"""

***Prompt for classification inference Predict diabetes tuned model 3***

ict_prompt_2 = """
Use the classification model:
/path/to/predict_diabetes/models/tuned_model_3/tuned_model_3.pkl
Make predictions using the predictors in the file: /path/to/predict_diabetes/test_set.csv
The ground truth values are in the column: "Outcome"

Output directory: "/path/to/ouput_directory"
Deploy the proper agent for this task.
"""

***Prompt for classification inference Predict diabetes blended model***

ict_prompt_3 = """
Use the classification model:
/path/to/predict_diabetes/models/blended_model/blended_model.pkl
Make predictions using the predictors in the file: /path/to/predict_diabetes/test_set.csv
The ground truth values are in the column: "Outcome"
Output directory: "/path/to/ouput_directory"
"""

***Prompt for classification inference Predict heart disease tuned model 3***

ict_prompt_4 = """
Use the classification model:
/path/to/heart_disease/models/tuned_model_3/tuned_model_3.pkl
Make predictions using the features in the file:
/path/to/heart_disease/independent_eval_cohort.csv
The ground truth values are in the column: "target"
Output directory: "/path/to/ouput_directory"
"""

***Prompt for classification inference Predict heart failure tuned model 2***

ict_prompt_5 = """
Use the classification model:
/path/to/heart_failure/models/tuned_model_2/tuned_model_2.pkl
Make predictions using the predictors in the file:
/path/to//heart_failure/independent_eval_cohort.csv
The ground truth values are in the column: "DEATH_EVENT"
Output directory: "/path/to/ouput_directory"
"""

**Training prompts Regression Model (Tabulated data)**

***Prompt for regression model development Life expectancy dataset***

trt_prompt_1 = """
Train a regression model using the file: /path/to/Life-Expectancy-Data.csv
Exclude "lightgbm".
Target column: "Life_expectancy".
Save the results here: "/path/to/ouput_directory"
"""

**Inference prompts Regression Model (Tabulated data)**

***Prompt for regression inference Life expectancy (providing the gt for comparison)***

irt_prompt_1 = """
Use the regression model:
/path/to/life_expectancy/models/tuned_model_1/tuned_model_1.pkl
Make predictions using the predictors in the file:
/path/to/life_expectancy/independent_eval_cohort.csv
The ground truth values are in the column: "Life_expectancy"
Output directory: /path/to/ouput_directory
Deploy the proper agent and tool for this task
"""

***Prompt for regression inference Life expectancy (without gt)***

irt_prompt_2 = """
Use the regression model:
/path/to/life_expectancy/models/tuned_model_1/tuned_model_1.pkl
Make predictions using the predictors in the file:
/path/to/life_expectancy/no_gt_independent_eval_cohort.csv
Output directory: /path/to/ouput_directory
"""

**Training Prompts - Image Classification**

***Prompt for classification resnet18 model development for pneumoniamnist_28 dataset***

ict_prompt_1 = """
Deploy the appropriate agent and tool to train a classification resnet18 model.
The train dataset directory: "/path/to/pneumoniamnist_28/dataset_pneumoniamnist_28/train",
the validation dataset directory:
"/path/to/pneumoniamnist_28/dataset_pneumoniamnist_28/val",
the test dataset directory: "/path/to/pneumoniamnist_28/dataset_pneumoniamnist_28/test",
Number of classes 2.
Use a batch size of 64. Number of epochs: 60
Output directory: "/path/to/ouput_directory"
"""

***Prompt for classification resnet34 model development for pathmnist_64 dataset***

ict_prompt_2 = """
Train a classification resnet34 model.
The train, validation and test datasets: /path/to/pathmnist_64/dataset_pathmnist_64.
Number of classes 9.
Use a batch size of 32. Set patience to 10 and number of epochs to 50.
Output directory: /path/to/ouput_directory
"""

***Prompt for classification resnet50 model development for breastmnist_128 dataset***

ict_prompt_3 = """
Train a classification resnet50 model.
The train, val and test data are available here:
"/path/to/breastmnist_128/dataset_breastmnist_128/".
Number of classes 2.
Train for 50 epochs. Do not use early stopping.
Output folder: "/path/to/ouput_directory"
"""

***Prompt for classification resnet101 model development for dermamnist_224 dataset***

ict_prompt_4 = """
Train a classification resnet101 model.
The train, val and test data are available here:
"/path/to/dermamnist_224/dataset_dermamnist_224".
Number of classes 7. Train for 200 epochs. Set patience for early stopping to 10.
Output folder: "/path/to/ouput_directory"
"""

***Prompt for classification resnet152 model development for organamnist_28 dataset***

ict_prompt_5 = """
Train a classification resnet152 model.
The train, val and test data are available here:
"/path/to/organamnist_28/dataset_organamnist_28/".
Number of classes 11. Set patience to 5.
Output folder: "/path/to/ouput_directory"
"""

***Prompt for classification vgg16 model development for octmnist_28 dataset***

ict_prompt_6 = """
Train a classification vgg16 model.
The train, val and test data are available here: "/path/to/octmnist_28/dataset_octmnist_28/".
Number of classes 4. Do not use pretrained weights.
Output folder: "/path/to/ouput_directory"
"""

***Prompt for classification InceptionV3 model development for bloodmnist_128 dataset***

ict_prompt_7 = """
Train a classification InceptionV3 model.
The train, val and test data are available here:
"/path/to/bloodmnist_128/dataset_bloodmnist_128".
Number of classes 8. Use pretrained weights and a batch size of 64. Train for 150 epochs.
Output folder: "/path/to/ouput_directory"
"""

**Inference Prompts - Image Classification**

***Prompt for inference with resnet18 model - pneumoniamnist_28 dataset***

ici_prompt_1 = """
Use the resnet-18 model available here:
"/path/to/pneumoniamnist_28/output_master_agent/resnet18_ict_prompt_1/best_model.pt",
to classify the images in this folder:
"/path/to/pneumoniamnist_28/dataset_pneumoniamnist_28/test".
The number of classes is 2.
The ground truth labels for the evaluation are availabe
here:"/path/to/pneumoniamnist_28/inference/gt_test_labels.csv".
Save the evaluation output in this directory: "/path/to/ouput_directory".
Deploy the appropriate agent and tool for this task.
"""

***Prompt for inference with resnet34 model - pathmnist_64 dataset***

ici_prompt_2 = """
Use the resnet34 model available here:
"/path/to/pathmnist_64/output_master_agent/resnet34_ict_prompt_2/best_model.pt",
to classify the images in this folder: "/path/to/pathmnist_64/dataset_pathmnist_64/test".
The number of classes is 9.
The ground truth labels for the evaluation are availabe
here:"/path/to/pathmnist_64/inference/gt_test_labels.csv".
Save the evaluation output in this directory: "/path/to/ouput_directory".
"""

***Prompt for inference with resnet50 model - breastmnist_128 dataset***

ici_prompt_3 = """
Use the resnet50 model available here:
"/path/to/breastmnist_128/output_master_agent/resnet50_ict_prompt_3/best_model.pt",
to classify the images in this folder: "/path/to/breastmnist_128/dataset_breastmnist_128/test".
The number of classes is 2.
The ground truth labels for the evaluation are availabe here:
"/path/to/breastmnist_128/inference/gt_test_labels.csv".
Output directory: "/path/to/ouput_directory".
"""

***Prompt for inference with resnet101 model - dermamnist_224 dataset***

ici_prompt_4 = """
Use the resnet101 model available here:
"/path/to/dermamnist_224/output_master_agent/resnet101_ict_prompt_4/best_model.pt",
to classify the images in this folder: "/path/to/dermamnist_224/dataset_dermamnist_224/test".
Number of classes 7.
Output folder: "/path/to/ouput_directory".

"""

***Prompt for inference with resnet152 model - organamnist_28 dataset***
ici_prompt_5 = """
Use the resnet152 model available here:
"/path/to/dermamnist_224/output_master_agent/resnet101_ict_prompt_4/best_model.pt",
to classify the images in this folder: "/path/to/dermamnist_224/inference_test/images".
Number of classes 11.
Output folder: "/path/to/ouput_directory".
"""

***Prompt for inference with vgg16 model - octmnist_28 dataset***
ici_prompt_6 = """
Use the vgg16 model available here:
"/path/to/octmnist_28/output_master_agent/vgg16_ict_prompt_6/best_model.pt",
to classify the images in this folder: "/path/to/octmnist_28/dataset_octmnist_28/test".
Number of classes 4. The ground truth labels are availabe in this directory:
"/path/to/octmnist_28/inference".
Output folder: "/path/to/ouput_directory".
"""

***Prompt for inference with InceptionV3 model - bloodmnist_128 dataset***
ici_prompt_7 = """
Use the InceptionV3 model available here:
"/path/to/bloodmnist_128/output_master_agent/InceptionV3_ict_prompt_7/best_model.pt",
to classify the images in this folder: "/path/to/bloodmnist_128/dataset_bloodmnist_128/test".
Number of classes 8. The ground truth labels are availabe in this directory:
"/path/to/bloodmnist_128/inference/gt_test_labels.csv".
Output folder: "/path/to/ouput_directory".
"""

**Multi-tasking prompts**

***First Multi-tasking Test Case***
multi_task_prompt_1 = """
PHASE 1: SPLEEN SEGMENTATION
 For each CT scan file in
 "tests/test_multi_tasking/multi_task_prompt_1/ct_scans":
 1. Run TotalSegmentator with the total task to segment only the spleen.
 2. Locate the generated spleen mask (typically named "spleen.nii.gz" in
 the output directory)
 3. Rename the spleen mask to match the original CT scan filename exactly:
 - Original: "1.nii.gz" → Mask: "1.nii.gz"
 - Original: "patient_001.nii.gz" → Mask: "patient_001.nii.gz"

4. Move the renamed mask to:
"tests/test_multi_tasking/multi_task_prompt_1/masks"
5. Clean up temporary segmentation files (keep only the renamed spleen mask)

PHASE 2: RADIOMIC FEATURE EXTRACTION
After ALL spleen masks are successfully created:
6. For each CT scan and its corresponding mask:
- Extract radiomic features
7. Combine all radiomic features into a single CSV file
- Include a column identifying the source CT scan filename
- Save as:
"tests/test_multi_tasking/multi_task_prompt_1/radiomic_features/radiomic_features.csv"

PHASE 3: EXPLORATORY DATA ANALYSIS
8. Load the radiomic features CSV file
9. Perform EDA (NO PLOTS)
10. Save all EDA results in: "tests/test_multi_tasking/multi_task_prompt_1/eda_results/"
"""

***Second Multi-tasking Test Case***

multi_task_prompt_2 = """
Using the nnUNet dataset 135, 3d full res, fold_all model, segment the mpMRI scans located in: "tests/test_multi_tasking/multi_task_prompt_2/mpMRI_scans".
Save the mask files to: "tests/test_multi_tasking/multi_task_prompt_2/masks".
After segmentation:
1)For each predicted mask file (e.g., BraTS2021_00000.nii.gz), rename the mask file so that it matches the corresponding T1 image filename.
For example:
If the mask is BraTS2021_00000.nii.gz and the corresponding T1 scan is BraTS2021_00000_0000.nii.gz, then rename the mask file to BraTS2021_00000_0000.nii.gz (i.e., copy the T1 filename exactly).
Once all mask files are correctly renamed:
2)Use the T1-weighted MRI scans available here: "tests/test_multi_tasking/multi_task_prompt_2/T1_scans" and their corresponding masks available here: "tests/test_multi_tasking/multi_task_prompt_2/masks"
to extract radiomic features for each label in the masks.
Use the following image filters: Exponential, Original, Wavelet.
3)Save each resulting radiomic features CSV file in: "tests/test_multi_tasking/multi_task_prompt_2/radiomic_features".
4)Perform Exploratory Data Analysis (EDA) for the extracted features separately for each label. Do not generate any plots.
Save the EDA results for each label in subfolders inside: "tests/test_multi_tasking/multi_task_prompt_2/eda_results".

"""

***Third Multi-tasking Test Case***

multi_task_prompt_3 = """
Perform a feature importance analysis for the file:
"/path/to/test_multi_tasking/multi_task_prompt_3/mamamia_features/dataset_mamamia.csv"
.
Save three csv files with the top 20, 50 and 100 features here: "/path/to/ouput_directory".
Targert column is "pcr".
Then use the csv file with the top 20 features to train a classifier. Save the output of the training process in this directory: "/path/to/ouput_directory".
"""

***Fourth Multi-tasking Test Case***

multi_task_prompt_4 = """
Perform exploratory data analysis for the file:
"/path/to/breast_cancer_wisconsin_diagnosis_dataset.csv".
Save eda results and a csv files with the top 10 features here: "/path/to/ouput_directory".
Targert column is "diagnosis".
Then use the csv file with the top 10 features to train a classifier. Save the output of the training process in this directory: "/path/to/ouput_directory".
"""

***Fifth Multi-tasking Test Case***

multi_task_prompt_5 = """
Train an InceptionV3 classification model using data available in the following directory:
"/path/to/pneumoniamnist_128/dataset_pneumoniamnist_128".
The number of classes is 2. Train for 100 epochs. Set patience to 10.
Then, use the trained model to classify images in this directory:
"/path/to/pneumoniamnist_128/inference/test_data".
The ground truth labels for the evaluation are availabe here:
"/path/to/pneumoniamnist_128/inference/gt_test_labels.csv".
Save the trained model and inference results in this directory: "/path/to/ouput_directory".
"""